\documentclass[]{spie}  

\usepackage{amsmath,amsfonts,amssymb}
\usepackage{graphicx}
\usepackage[colorlinks=true, allcolors=blue]{hyperref}
\usepackage{amssymb}
\usepackage{adjustbox}
\usepackage{graphicx}
\usepackage{siunitx}
\usepackage{mathtools}
\usepackage{threeparttable}
\usepackage{float}
\usepackage{booktabs}
\usepackage{tabularx}
\usepackage{makecell}
\newcolumntype{C}[1]{>{\centering\arraybackslash}p{#1}}
\newcolumntype{L}[1]{>{\raggedright\arraybackslash}p{#1}}
\usepackage{multicol}
\usepackage[T1]{fontenc}
\usepackage{csquotes}
\usepackage{url}
\usepackage{subcaption}
\usepackage{hhline}
\usepackage{textcomp}
\usepackage{multirow}
\usepackage{makecell}

\usepackage{siunitx}
\newcommand\percentage[2][round-precision = 2]{
    \SI[round-mode = places,
        scientific-notation = fixed, fixed-exponent = 0,
        output-decimal-marker={.}, #1]{#2e2}{\percent}%
}

\graphicspath{ 
{fig_money/} 
{fig_results/}
}

\title{Correcting Class Imbalances with Self-Training for \\
Improved Universal Lesion Detection and Tagging}

\author[1]{Alexander Shieh}
\author[1]{Tejas Sudharshan Mathai}
\author[1]{Jianfei Liu}
\author[2]{\\Angshuman Paul}
\author[1]{Ronald M. Summers}
\affil[1]{Imaging Biomarkers and Computer-Aided Diagnosis Laboratory, Radiology and Imaging Sciences, Clinical Center, National Institutes of Health, Bethesda MD, USA}
\affil[2]{Indian Institute of Technology, Jodhpur, Rajasthan, India}

\authorinfo{Send correspondence to T.S.M.: tejas dot mathai at nih dot gov}

\begin{document} 
\maketitle

\begin{abstract}

Universal lesion detection and tagging (ULDT) in CT studies is critical for tumor burden assessment and tracking the progression of lesion status (growth/shrinkage) over time. However, a lack of fully annotated data hinders the development of effective ULDT approaches. Prior work used the DeepLesion dataset (4,427 patients, 10,594 studies, 32,120 CT slices, 32,735 lesions, 8 body part labels) for algorithmic development, but this dataset is not completely annotated and contains class imbalances. To address these issues, in this work, we developed a self-training pipeline for ULDT. A VFNet model was trained on a limited 11.5\% subset of DeepLesion (bounding boxes + tags) to detect and classify lesions in CT studies. Then, it identified and incorporated novel lesion candidates from a larger unseen data subset into its training set, and self-trained itself over multiple rounds. Multiple self-training experiments were conducted with different threshold policies to select predicted lesions with higher quality and cover the class imbalances. We discovered that direct self-training improved the sensitivities of over-represented lesion classes at the expense of under-represented classes. However, upsampling the lesions mined during self-training along with a variable threshold policy yielded a 6.5\% increase in sensitivity at 4 FP in contrast to self-training without class balancing (72\% vs 78.5\%) and a 11.7\% increase compared to the same self-training policy without upsampling (66.8\% vs 78.5\%). Furthermore, we show that our results either improved or maintained the sensitivity at 4FP for all 8 lesion classes.

\end{abstract}





\section{Introduction}
\label{intro}  

Localizing metastatic lesions is critical for the assessment of tumor burden and lesion status change (growth, shrinkage, unchanged), and it determines the course of patient therapy. Presently, computed tomography (CT) is the preferred modality for imaging patients \cite{Nishino2010RECIST}, but lesions in CT can have diverse shapes and appearances with some occurring in standard locations (e.g. liver or lung) while others are rarer (e.g. bone). Radiologists rely on the lesion size as an indicator of malignancy, and size them using RECIST guidelines \cite{Nishino2010RECIST}. However, this guideline varies across institutions due to many factors, such as clinical practice, type of CT scanners, exam protocols, contrast phases used. Presently, automated universal lesion detection (ULD) approaches \cite{Yan2018_deeplesion,Tarun2022ensemble,Yang2020_alignShift,Cai2021_LesionHarvester,Yan2021_LENS,Tang2019_uldor,Xie2021_recistnet,Zlocha2019,Yan2018_3dce,Yang2021_A3D,Li2022_SATR} identify all lesions in a given CT study. ULD with tagging (ULDT) adds diagnostic value by tagging the lesion with the body part in which the lesion is located. Prior work on ULDT \cite{Yan2019_LesaNet,Yan2019_MULAN,Erickson2022_classImbalanceCorrection,Naga2022_weakSupSelfTraining} used the publicly available DeepLesion dataset \cite{Yan2018_deeplesion}, but it is incomplete as only clinically significant lesions are annotated while others remain unannotated. It is also severely class-imbalanced \cite{Erickson2022_classImbalanceCorrection} with considerable over-representation of certain classes (e.g. liver, lung) over other under-represented classes (e.g. bone, kidney). Moreover, lesion tags are only available for the DeepLesion validation and testing splits (30\%). 

\begin{figure}[!t]
\centering
\begin{subfigure}[t]{0.35\columnwidth}
\vspace*{\fill}
  \centering
  \includegraphics[width=\columnwidth,height=3.25cm]{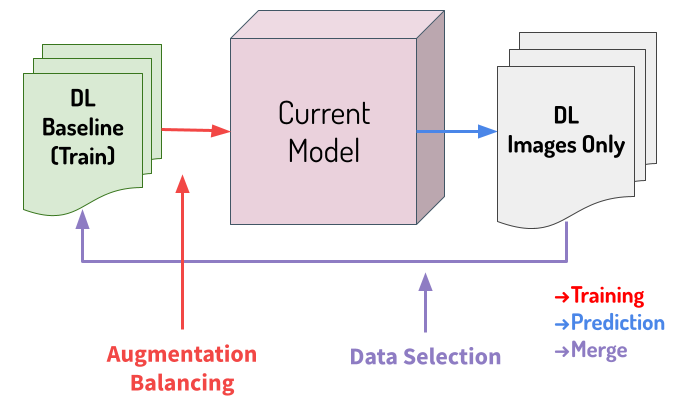}
  \centerline{(a)} 
\end{subfigure} 
\begin{subfigure}[t]{0.2\columnwidth}
\vspace*{\fill}
  \centering
  \includegraphics[width=\columnwidth,height=3.25cm]{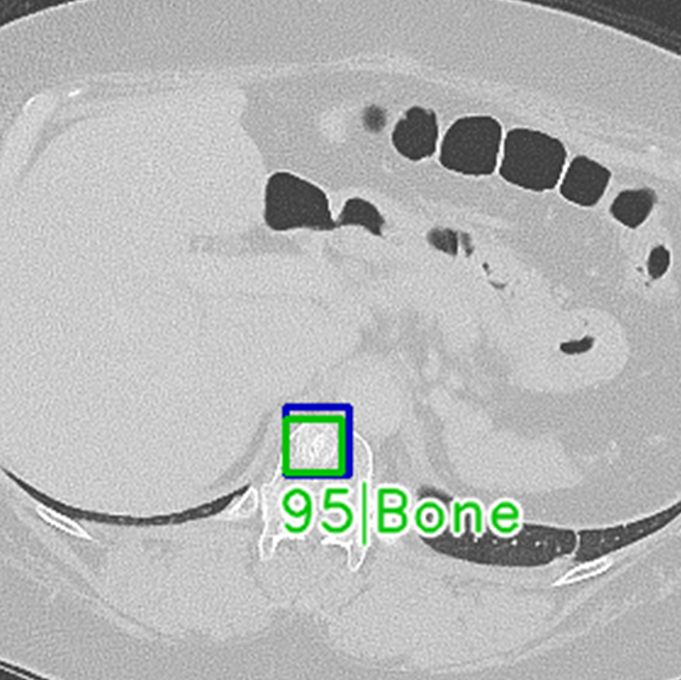}
  \centerline{(b)} 
\end{subfigure} 
\begin{subfigure}[t]{0.2\columnwidth}
\vspace*{\fill}
  \centering
  \includegraphics[width=\columnwidth,height=3.25cm]{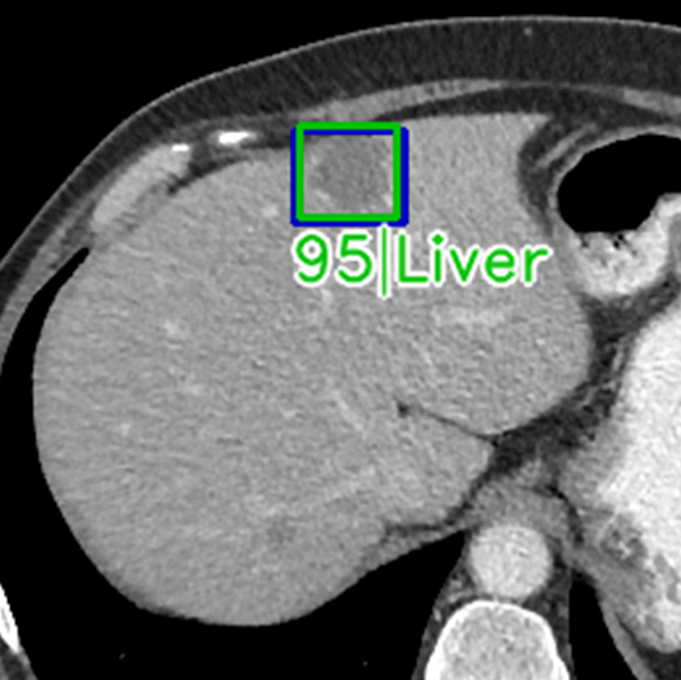}
  \centerline{(c)}
\end{subfigure} 
\begin{subfigure}[t]{0.2\columnwidth}
\vspace*{\fill}
  \centering
  \includegraphics[width=\columnwidth,height=3.25cm]{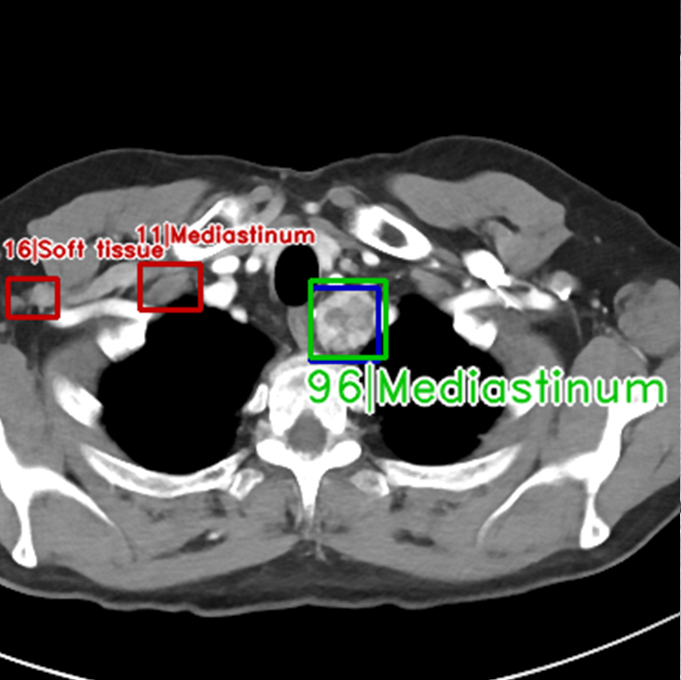}
  \centerline{(d)}
\end{subfigure} 
\caption{\label{fig:money} (a) Self-training pipeline. ``DL'' stands for DeepLesion dataset. (b) A mined ``bone'' lesion of the thoracic vertebra. (c) A mined hypodense ``liver'' lesion. (d) A mined ``mediastinum'' lesion with 2 low-confidence FP also visible. (b)-(d) represent lesion mining results on the original DeepLesion training split. The green boxes are lesions that were found during self-training. The red boxes are FP with confidence $<$ 90\% and were not selected during self-training. The blue boxes are original annotations provided by DeepLesion. They are only plotted for visualization, and were discarded during self-training. }
\label{fig:Self-Train}
\end{figure}

To leverage the incomplete annotations of DeepLesion and improve the dataset, we propose a self-training pipeline for ULDT. Our approach used a limited 11.5\% data subset of DeepLesion (bounding boxes + tags) to train a VFNet model\cite{Zhang2021_vfnet} for lesion detection and tagging. Subsequently, the model predicted new lesions on a larger data subset (DeepLesion training split), incorporated those that crossed a confidence threshold, and integrated them into the dataset it was originally trained on. In this manner, VFNet learned from its own predictions and efficiently re-trained itself over multiple rounds. The final model was used for lesion detection and tagging, and an overview of the process is shown in Fig \ref{fig:money}. Various experiments were conducted with different threshold policies to select mined lesion predictions for inclusion in the training dataset to re-train the model. To the best of our knowledge, we are the first to employ self-training in a fully unsupervised manner, upsample the dataset of mined lesions to balance the quantities for different lesion classes, and investigate both their effects for the task of ULDT.

\section{Methods}

\noindent
\textbf{Dataset.} The DeepLesion dataset had 32,735 lesions prospectively annotated with 2D bounding boxes on 32,120 CT slices. Context slices 30 mm above and below the annotated slice were also provided, but they were not annotated. DeepLesion was divided into official training (${O}_{Tr}$, 70\%), validation ($O_{V}$, 15\%) and testing ($O_{T}$, 15\%) splits with the lesions in the validation and testing splits tagged with one of 8 body part labels (bone, abdomen, mediastinum, liver, lung, kidney, soft tissue, pelvis). Patients from $O_{Tr}$ were used solely for unsupervised self-training and their annotations (boxes with no tags) were discarded. We used the slice overlap between a fully annotated test set (boxes with no tags) described in prior work\cite{Yan2021_LENS} and the DeepLesion test set as our final testing set $F_{T}$ throughout this study. This step provided a near-completely annotated test set with tags mapped from DeepLesion. We then removed the patients in $F_{T}$ from $O_{V} \cup O_{T}$ to ensure no patient-level overlap. Following this, we randomly sampled patients from $O_{V} \cup O_{T}$, and sub-divided them into our final training $F_{Tr}$ (70\%) and validation $F_{V}$ (30\%) splits. The resultant $F_{Tr}$ contained only 11.5\% of the total number of lesions in the original DeepLesion dataset. All splits are done at patient level and no patient overlaps exists between $F_{Tr}$, $F_{V}$ and $F_{T}$ (see supplementary material Fig. \ref{fig:Data_Diagram} and Table \ref{tab:dataset} for the construction of these subsets).

\noindent
\textbf{Model.} The VarifocalNet (VFNet) model \cite{zhang2020varifocalnet} was used in this work as it was shown to be very effective for detection of lesions \cite{Tarun2022ensemble}. For more details on the model, we refer the reader to prior work \cite{zhang2020varifocalnet,Tarun2022ensemble}. VFNet was trained on $F_{Tr}$, and its predictions were combined using weighted boxes fusion (WBF) \cite{solovyev2021weighted}. For the sake of brevity, model implementation details are provided in Sec. \ref{suppMat_modelImplementation} of the supplementary material. 

\noindent
\textbf{Self-Training with Intra- and Inter-patient Mined Lesions.} Once the baseline VFNet model (Round 0) was trained, it predicted bounding boxes for new lesions in slices in $F_{Tr}$ (intra-patient data) and $O_{Tr}$ (inter-patient data). The predictions were filtered by their confidences based on a threshold policy (see experimental design in Sec. \ref{sec:exp_results}). Slices from $O_{Tr}$ without predictions exceeding the confidence threshold were discarded resulting in a new dataset $O^{\prime}_{Tr}$, which was merged with $F_{Tr}$ to form the new self-training set ($F^{\prime}_{Tr} \leftarrow F_{Tr} \cup O^{\prime}_{Tr}$) to train the model (from scratch) for ensuing rounds. This process was repeated for a total of 4 mining rounds. 

\noindent
\textbf{Dataset Upsampling.} During each mining round, upsampling of lesion quantities was also performed. First, the tag with the most predicted lesions (e.g. ``lung'') in $F^{\prime}_{Tr}$ was found, and the lesions from all other classes were upsampled to match the quantity of the prevalent class. Concretely, if 1000 lesions were predicted for class $C_1$ while a different class $C_2$ had 200 lesions, then $C_2$ was repeated 5 times in $F^{\prime}_{Tr}$ to match the quantity of $C_1$.

\section{Experiments and Results}
\label{sec:exp_results}

\textbf{Experimental Design.} 
We designed three experiments to filter predicted lesions by their confidences. In the first experiment $E_{S}$, a static confidence threshold $T_{S}$ of 90\% and a 50\% box IoU overlap was employed. Our hypothesis was that high quality lesions would be mined across rounds with this policy. Predicted lesions with confidences $\geq T_{S}$ and IoU overlaps $\geq$ 50\% were incorporated into $F^{\prime}_{Tr}$ for ensuing mining rounds. In the second experiment $E_{SV}$, a semi-variable confidence threshold policy starting at 90\% for 2 rounds followed by 85\% for two rounds was designed. Our assumption was that once high quality lesions were exhausted, dropping the threshold slightly would yield similar or better performance. In our last experiment $E_{V}$, we established a variable confidence threshold policy. A high confidence threshold of 90\% was set for the first round, and it was dropped by 5\% for subsequent rounds. Our aim was to mine good quality lesions in the first round, and harvest larger lesion quantities over the following rounds. For baseline comparisons, we compared the results of these experiments against one where the model underwent no self-training and one where the lesion upsampling was not used during self-training. Our sensitivity results at 4FP for a 50\% box IoU overlap are shown in Tables \ref{tab:ExpVFNetUpsampled}, \ref{tab:ExpVFNetNotUpsampled} and \ref{tab:ExpVFNetUpsampledfull}. We also compared the VFNet model used in this work against the YOLOX \cite{GeYOLOX2021} architecture, but noticed that YOLOX showed an inferior detection performance (see Table \ref{tab:ExpYOLOXUpsampled}). A contemporary transformer-based Deformable DETR \cite{Zhu2021DeformableDETR} model was also tested, but despite extensive hyper-parameter tuning, a stable performance was unattainable. 


\noindent
\textbf{Results - Self-Training \textit{with} data upsampling.} In Table \ref{tab:ExpVFNetUpsampled}, we contrasted the different threshold policies employed during self-training. The static threshold policy $E_{S}$ achieved 76.2\% sensitivity at 4FP, the semi-variable threshold policy $E_{SV}$ achieved 76.9\%, and the variable threshold policy $E_{V}$ achieved 78.5\%. The ensemble of models over 5 rounds with $E_{V}$ achieved the best detection sensitivity. Compared against the ensemble from $E_{S}$, 7/8 classes maintained or improved their sensitivity at 4FP (except ``lung'' class). Similarly, compared against the ensemble from $E_{SV}$, 7/8 classes had their sensitivity at 4FP improved or maintained (except ``abdomen'' class). The $E_{V}$ policy was most beneficial for underrepresented lesion types, such as ``bone'', ``kidney'', ``soft tissue'' and ``pelvis''. Comparing the self-trained model that used $E_{V}$ with dataset upsampling against the same self-trained model \textit{without dataset upsampling}, we observed that the former showed improvements for all 8/8 lesion classes. Interestingly, the distribution of the number of lesions mined is similar to the distribution at round 0 (baseline model without self-training). The sensitivity results for all rounds are shown in Supplementary Tables ~\ref{tab:ExpVFNetUpsampledfull} and ~\ref{tab:ExpVFNetNotUpsampled}.



\begin{table}[!h]
\centering\fontsize{10}{12}\selectfont 
\setlength\aboverulesep{0pt}\setlength\belowrulesep{0pt} 
\setlength{\tabcolsep}{7pt} 
\setcellgapes{3pt}\makegapedcells 
\caption{Results of the VFNet model self-trained with dataset upsampling for different threshold policies. Sensitivites are shown at 4 FP and at 50\% IoU threshold.}
\label{tab:ExpVFNetUpsampled}
\begin{center}
\begin{adjustbox}{max width=\textwidth, max height=4cm}
\begin{tabular}{@{} c|c|c|c|c|c|c|c|c|c @{}} 
\toprule
Round         & Bone       & Kidney     & Soft Tissue       & Pelvis       & Liver       & Mediastinum       & Abdomen       & Lung      & Mean\\
\midrule
\multicolumn{10}{c}{\textbf{No Self-Training (Baseline)}}\\
\midrule
Round 0      & 61.3\% & 62.6\% & 77.1\% & 69.8\% & 77.1\% & 73.9\% & 71.2\% & 82.8\% & 72.0\% \\
\midrule
\makecell{\# Lesions used}  & \makecell{97\\(2.6\%)} & \makecell{195\\(5.2\%)} & \makecell{288\\(7.6\%)} & \makecell{321\\(8.5\%)} & \makecell{426\\(11.3\%)} & \makecell{613\\(16.8\%)} & \makecell{788\\(20.9\%)} & \makecell{1039\\(27.6\%)} & - \\
\midrule
\multicolumn{10}{c}{\textbf{Static Threshold Policy [90\%, 90\%, 90\%, 90\%] }}\\
\midrule
Ensemble of Rounds            & 67.7\% & 72.2\% & 79.0\% & 76.0\% & 80.5\% & 77.5\% & 72.0\% & 84.6\% & 76.2\% \\
\midrule
\makecell{\# Lesions mined}  & \makecell{187\\(3.0\%)} & \makecell{539\\(8.5\%)} & \makecell{682\\(10.8\%)} & \makecell{761\\(12.0\%)} & \makecell{1142\\(18.1\%)} & \makecell{693\\(11.0\%)} & \makecell{768\\(12.2\%)} & \makecell{1545\\(24.5\%)} & - \\
\midrule
\multicolumn{10}{c}{\textbf{Semi-Variable Threshold Policy [90\%, 90\%, 85\%, 85\%]}}\\
\midrule
Ensemble of Rounds             & 71.0\% & 74.8\% & 81.0\% & 75.0\% & 81.0\% & 77.5\% & 72.6\% & 82.3\% &  76.9\% \\
\midrule
\makecell{\# Lesions mined}  & \makecell{301\\(2.7\%)} & \makecell{729\\(6.6\%)} & \makecell{1043\\(9.5\%)} & \makecell{1119\\(10.1\%)} & \makecell{1902\\(17.2\%)} & \makecell{1932\\(17.5\%)} & \makecell{1509\\(13.7\%)} & \makecell{2494\\(22.6\%)} & - \\
\midrule
\multicolumn{10}{c}{\textbf{Variable Threshold Policy [90\%, 85\%, 80\%, 75\%] }}\\
\midrule
Ensemble of Rounds             & 77.4\% & 76.5\% & 83.8\% & 76.0\% & 81.8\% & 78.3\% & 72.0\% & 82.4\% & 78.5\% \\
\midrule
\makecell{\# Lesions mined}  & \makecell{481\\(2.7\%)} & \makecell{1239\\(6.8\%)} & \makecell{1768\\(9.8\%)} & \makecell{1895\\(10.5\%)} & \makecell{3022\\(16.7\%)} & \makecell{3065\\(16.9\%)} & \makecell{2630\\(14.5\%)} & \makecell{4006\\(22.1\%)} & - \\
\bottomrule
\end{tabular}
\end{adjustbox}
\end{center}
\end{table}

\noindent
\textbf{Results - Self-Training \textit{without} data upsampling.} From the Supplementary Table ~\ref{tab:ExpVFNetNotUpsampled}, we saw the inconsistent detection sensitivities for different threshold policies. Specifically, the ensemble of the 5 mining rounds for the semi-variable threshold policy $E_{SV}$ performed the best with an average sensitivity of 67.6\% at 4FP in comparison to 64.9\% from the static threshold $E_{S}$ policy and 66.8\% from the variable threshold policy $E_{V}$ respectively. Only 6/8 lesion classes showed moderate sensitivity improvements (except ``bone'' and ``pelvis''). Although the results from the threshold policies are higher than that of the model that was not self-trained (round 0, baseline), they are still $\sim$9\% lower in contrast to the model trained with data upsampling for the same $E_{SV}$ policy (67.6\% vs 76.9\%). We believe that this was due to the highly imbalanced distribution of mined lesions as depicted in Table ~\ref{tab:ExpVFNetNotUpsampled}, thereby attesting to our strategy of upsampling mined lesions during training to balance them across all lesion classes. 


\noindent
\textbf{Results - Effect of data upsampling.} As seen in Tables \ref{tab:ExpVFNetUpsampled} and \ref{tab:ExpVFNetNotUpsampled}, the baseline model trained with upsampled data (Round 0 with no self-training) surpassed the baseline model without upsampling (Round 0 with no self-training) by $\sim6$\% mean sensitivity at 4 FP (72.0\% vs 65.4\%). All the lesion classes achieved a higher sensitivity at 4FP with upsampling in contrast to without upsampling. 



\section{Discussion and Conlcusion}

In this study, we explored the effectiveness of self-training for ULDT. To the best of our knowledge, we are the first to employ self-training in a fully unsupervised manner, upsample the dataset of mined lesions to balance the quantities for different lesion classes, and investigate the effects of both for ULDT in the DeepLesion dataset. We showed that self-training can indeed improve the detection sensitivity at 4 FP. However, straightforward self-training with the various threshold policies we tested tended to favor over-represented lesion types (e.g. lung, liver). The model was more inclined to increase the quantities of common or easier lesions at the expense of rarer lesion classes (e.g. bone, pelvis) since a high confidence threshold was used. Thus, only those lesion types improved by a larger margin. This was a major issue as both common and rare lesions should be detected with high sensitivity. We have shown that this can be corrected with balancing the training data via upsampling the mined lesions in the dataset. We do not discard any lesions that were mined during this lesion harvesting process. Consequently, our VFNet model self-trained with a variable threshold policy and with dataset upsampling achieved a 6.5\% increase in sensitivity at 4 FP in contrast to the baseline model without self-training (72\% vs 78.5\%) and a 11.7\% increase compared to the same self-training policy without dataset upsampling (66.8\% vs 78.5\%). We also compared VFNet's detection performance against YOLOX self-trained with dataset upsampling, but those results were far inferior as seen in Table ~\ref{tab:ExpYOLOXUpsampled}. 

In terms of limitations of our work, employing the same threshold during data selection process for all lesion types at each mining round may not be ideal \cite{Li2020Selective,ZHANG2022260automaticlearning}. As rare lesion classes (e.g. bone) tend to be predicted with lower confidences than common classes (e.g. lung, liver), a dynamic threshold (potentially rank-based) that adjusts to the lesion class may be warranted, and further investigations are needed. In addition, our search for the best threshold strategy over multiple self-training rounds was not exhaustive and should be explored in future work.
We also noted that predicted lesions with high confidence scores might not accurately correlate with the ground truth, and other semi-supervised methods can be explored for improved data selection \cite{Li2020Selective}. In this work, we strive towards a fundamental understanding of these research thrusts in order to leverage the vast amount of unannotated data and unleash the full potential of ULDT in the clinical workflow.

\section{Acknowledgements} 

This work was supported by the Intramural Research Program of the NIH Clinical Center. 

\clearpage
\bibliography{report} 
\bibliographystyle{spiebib} 

\clearpage

\section{SUPPLEMENTARY MATERIAL}

   \begin{figure} [ht]
   \begin{center}
   \begin{tabular}{c} 
   \includegraphics[height=9cm]{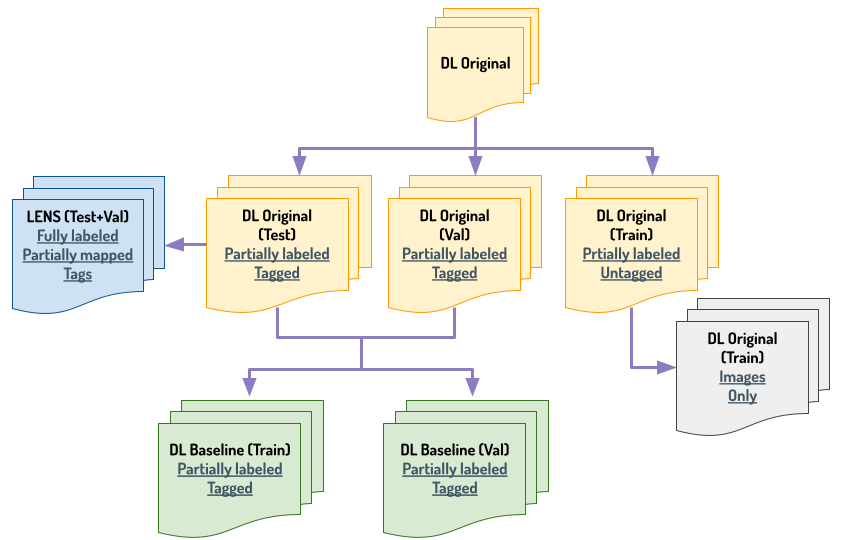}
   \end{tabular}
   \end{center}
   \caption[example] 
   { \label{fig:Data_Diagram} The construction of our training, testing and validation splits.}
   \end{figure} 

\subsection{Model Training and Implementation}
\label{suppMat_modelImplementation}
The input for the model was the annotated slice along with 2 adjacent slices before and after the annotated slice. For upsampling, we computed the number of predicted lesions for the most predominant class in the current training set, and then upsampled the number of lesions of all other classes to match that of the prevalent class. Despite the occurrence of multiple lesions with different classes on the same annotated slice, we still added the lesions from that slice regardless of whether the target lesion number was reached in order to reduce false negatives. As such, a very slight disparity in the final number of upsampled lesions per class was seen during the upsampling process (difference of $\leq$ 20 lesions in total upsampled). For data augmentation, we used combinations of random flips, resizing, cropping and rotations. We trained the model for a maximum of 30 epochs. 

During testing, for each round of self-training, we selected the weights from 5 epochs with the lowest validation loss to form an ensemble for prediction. However, the performance for individual lesion types fluctuated between different rounds as seen in Tables \Ref{tab:ExpVFNetNotUpsampled}, \ref{tab:ExpVFNetUpsampledfull}, and \Ref{tab:ExpYOLOXUpsampled}. To better assess the effect of the entire data selection policy, we evaluated another 5-model ensemble consisting of only the best epoch with the lowest validation loss from each round (Round 0 to Round 4) of the same threshold policy. The testing intersection over union (IoU) threshold between ground truth and predicted bounding boxes was 0.5\cite{Yan2019_MULAN}. The models were implemented using the mmDetection library and trained using a single NVIDIA Tesla V100 SXM2 32 GB GPU. The stochastic gradient descent algorithm was used for optimization, and the learning rate was 0.0001 with momentum 0.9. Batch size was fixed to 2 for all experiments.


\begin{table}[ht]
\caption{The number of patients, studies, series, slices and lesions of the original DeepLesion dataset, and the corresponding numbers used in our self-training, training, validation and testing splits. Note that since we performed patient level splits, the number of patients in $O_{Tr}$, $F_{T}$, $F_{Tr}$ and $F_{V}$ adds up to 4,427. We discarded the lesions in $O_{Tr}$ and used the images only.} 
\label{tab:dataset}
\begin{center} 
\def\arraystretch{1.5}
\begin{tabular}{|l|c|c|c|c|c|}
\hline
                       & Patients & Studies & Series & Slices & Lesions \\ \hhline{|=|=|=|=|=|=|}
$O$            & 4427     & 10594   & 14601  & 32120  & 32735   \\ \hline
$O_{Tr}$    & 3059     & 7386    & 10224  & 22496  & 22919   \\ \hline
$F_{T}$         & 586      & 771     & 847    & 1223   & 1416    \\ \hline
$F_{Tr}$    & 547      & 1231    & 1687   & 3695   & 3767    \\ \hline
$F_{V}$    & 235      & 524     & 693    & 1514   & 1548    \\ \hline
\end{tabular}
\end{center}
\end{table}

   \begin{figure} [ht]
   \begin{center}
   \begin{tabular}{c} 
   \includegraphics[height=5cm]{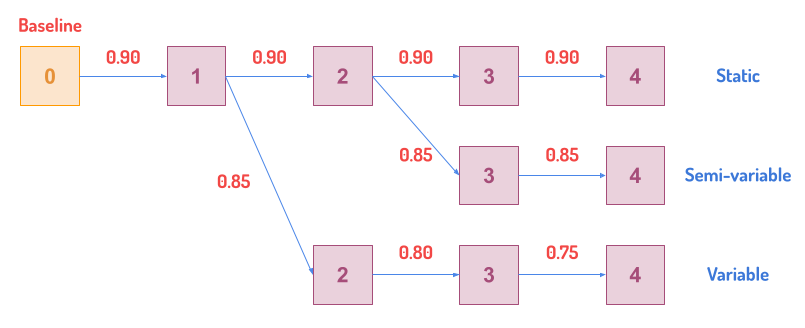}
   \end{tabular}
   \end{center}
   \caption[example] 
   { \label{fig:threshold} The three threshold policies used for our data selection process.}
   \end{figure} 

\begin{table}[ht]
\centering\fontsize{10}{12}\selectfont 
\setlength\aboverulesep{0pt}\setlength\belowrulesep{0pt} 
\setlength{\tabcolsep}{7pt} 
\setcellgapes{3pt}\makegapedcells 
\caption{Results of the VFNet model self-trained \textit{without dataset upsampling} for different threshold policies. Sensitivities are shown at 4 FP and at 50\% IoU threshold.}
\label{tab:ExpVFNetNotUpsampled}
\begin{center}
\begin{adjustbox}{max width=\textwidth, max height=7cm}
\begin{tabular}{@{} c|c|c|c|c|c|c|c|c|c @{}} 
\toprule
Round         & Bone       & Kidney     & Soft Tissue       & Pelvis       & Liver       & Mediastinum       & Abdomen       & Lung      & Mean (95\% CI)\\
\midrule
\multicolumn{10}{c}{\textbf{No Self-Training (Baseline)}}\\
\midrule
Round 0      & \percentage{0.581} & \percentage{0.548} & \percentage{0.667} & \percentage{0.688} & \percentage{0.699} & \percentage{0.672} & \percentage{0.579} & \percentage{0.799} &  \percentage{0.654} \\
\midrule
\makecell{Lesions\\at baseline}  & \makecell{97\\(2.6\%)} & \makecell{195\\(5.2\%)} & \makecell{288\\(7.6\%)} & \makecell{321\\(8.5\%)} & \makecell{426\\(11.3\%)} & \makecell{613\\(16.8\%)} & \makecell{788\\(20.9\%)} & \makecell{1039\\(27.6\%)} & - \\
\midrule
\multicolumn{10}{c}{\textbf{Static Policy [90\%, 90\%, 90\%, 90\%] }}\\
\midrule
Round 1 (90\%)        & \percentage{0.484} & \percentage{0.574} & \percentage{0.686} & \percentage{0.635} & \percentage{0.720} & \percentage{0.640} & \percentage{0.588} & \percentage{0.803} &  \percentage{0.641} \\
Round 2 (90\%)        & \percentage{0.516} & \percentage{0.565} & \percentage{0.686} & \percentage{0.625} & \percentage{0.750} & \percentage{0.664} & \percentage{0.625} & \percentage{0.818} & \percentage{0.656} \\
Round 3 (90\%)        & \percentage{0.419} & \percentage{0.583} & \percentage{0.657} & \percentage{0.635} & \percentage{0.733} & \percentage{0.609} & \percentage{0.588} & \percentage{0.790} &  \percentage{0.627} \\
Round 4 (90\%)        & \percentage{0.516} & \percentage{0.530} & \percentage{0.686} & \percentage{0.625} & \percentage{0.763} & \percentage{0.644} & \percentage{0.617} & \percentage{0.799} &  \percentage{0.648} \\
Round 0-4             & \percentage{0.484} & \percentage{0.548} & \percentage{0.705} & \percentage{0.625} & \percentage{0.733} & \percentage{0.668} & \percentage{0.617} & \percentage{0.815} & \percentage{0.649} \\
\midrule
\makecell{Lesions mined\\for Round 4}  & \makecell{2\\(0.1\%)} & \makecell{0\\(0.0\%)} & \makecell{53\\(3.1\%)} & \makecell{5\\(0.3\%)} & \makecell{82\\(4.7\%)} & \makecell{324\\(18.7\%)} & \makecell{22\\(1.3\%)} & \makecell{1242\\(71.8\%)} & - \\
\midrule
\multicolumn{10}{c}{\textbf{Semi-variable Policy [90\%, 90\%, 85\%, 85\%]}}\\
\midrule
Round 1 (90\%)        & \percentage{0.484} & \percentage{0.574} & \percentage{0.686} & \percentage{0.635} & \percentage{0.720} & \percentage{0.640} & \percentage{0.588} & \percentage{0.803} & \percentage{0.641} \\
Round 2 (90\%)        & \percentage{0.516} & \percentage{0.565} & \percentage{0.686} & \percentage{0.625} & \percentage{0.750} & \percentage{0.664} & \percentage{0.625} & \percentage{0.818} & \percentage{0.656} \\
Round 3 (85\%)        & \percentage{0.419} & \percentage{0.513} & \percentage{0.638} & \percentage{0.667} & \percentage{0.720} & \percentage{0.592} & \percentage{0.568} & \percentage{0.784} & \percentage{0.613} \\
Round 4 (85\%)        & \percentage{0.452} & \percentage{0.556} & \percentage{0.648} & \percentage{0.635} & \percentage{0.750} & \percentage{0.625} & \percentage{0.602} & \percentage{0.799} & \percentage{0.633} \\
Round 0-4             & \percentage{0.516} & \percentage{0.583} & \percentage{0.733} & \percentage{0.677} & \percentage{0.763} & \percentage{0.684} & \percentage{0.631} & \percentage{0.821} & \percentage{0.676} \\
\midrule
\makecell{Lesions mined\\for Round 4}  & \makecell{7\\(0.1\%)} & \makecell{7\\(0.1\%)} & \makecell{323\\(6.1\%)} & \makecell{144\\(2.7\%)} & \makecell{746\\(14.1\%)} & \makecell{1201\\(22.7\%)} & \makecell{467\\(8.8\%)} & \makecell{2389\\(45.2\%)} & - \\
\midrule
\multicolumn{10}{c}{\textbf{Variable Policy [90\%, 85\%, 80\%, 75\%] }}\\
\midrule
Round 1 (90\%)        & \percentage{0.484} & \percentage{0.574} & \percentage{0.686} & \percentage{0.635} & \percentage{0.720} & \percentage{0.640} & \percentage{0.588} & \percentage{0.803} & \percentage{0.641} \\
Round 2 (85\%)        & \percentage{0.613} & \percentage{0.565} & \percentage{0.638} & \percentage{0.625} & \percentage{0.733} & \percentage{0.684} & \percentage{0.588} & \percentage{0.796} & \percentage{0.655} \\
Round 3 (80\%)        & \percentage{0.419} & \percentage{0.591} & \percentage{0.648} & \percentage{0.719} & \percentage{0.780} & \percentage{0.688} & \percentage{0.631} & \percentage{0.780} & \percentage{0.657} \\
Round 4 (75\%)        & \percentage{0.516} & \percentage{0.557} & \percentage{0.705} & \percentage{0.667} & \percentage{0.767} & \percentage{0.680} & \percentage{0.666} & \percentage{0.799} & \percentage{0.670} \\
Round 0-4             & \percentage{0.484} & \percentage{0.617} & \percentage{0.714} & \percentage{0.667} & \percentage{0.763} & \percentage{0.668} & \percentage{0.611} & \percentage{0.821} & \percentage{0.668} \\
\midrule
\makecell{Lesions\\at baseline}  & \makecell{73\\(0.6\%)} & \makecell{200\\(1.7\%)} & \makecell{784\\(6.5\%)} & \makecell{836\\(7.0\%)} & \makecell{1678\\(14.0\%)}  & \makecell{2351\\(19.6\%)} & \makecell{2060\\(17.1\%)} & \makecell{4040\\(33.6\%)} & - \\
\bottomrule
\end{tabular}
\end{adjustbox}
\end{center}
\end{table}

\begin{table}[ht]
\centering\fontsize{10}{12}\selectfont 
\setlength\aboverulesep{0pt}\setlength\belowrulesep{0pt} 
\setlength{\tabcolsep}{7pt} 
\setcellgapes{3pt}\makegapedcells 
\caption{Results of the VFNet model self-trained with dataset upsampling for different threshold policies. Sensitivities are shown at 4 FP and at 50\% IoU threshold.}
\label{tab:ExpVFNetUpsampledfull} 
\begin{center}
\begin{adjustbox}{max width=\textwidth, max height=7cm}
\begin{tabular}{@{} c|c|c|c|c|c|c|c|c|c @{}} 
\toprule
Round         & Bone       & Kidney     & Soft Tissue       & Pelvis       & Liver       & Mediastinum       & Abdomen       & Lung      & Mean (95\% CI)\\
\midrule
\multicolumn{10}{c}{\textbf{No Self-Training (Baseline)}}\\
\midrule
Round 0      & \percentage{0.613} & \percentage{0.626} & \percentage{0.771} & \percentage{0.698} & \percentage{0.771} & \percentage{0.739} & \percentage{0.712} & \percentage{0.828} & \percentage{0.720} \\
\midrule
\makecell{Lesions\\at baseline}  & \makecell{97\\(2.6\%)} & \makecell{195\\(5.2\%)} & \makecell{288\\(7.6\%)} & \makecell{321\\(8.5\%)} & \makecell{426\\(11.3\%)} & \makecell{613\\(16.8\%)} & \makecell{788\\(20.9\%)} & \makecell{1039\\(27.6\%)} & - \\
\midrule
\multicolumn{10}{c}{\textbf{Static Policy [90\%, 90\%, 90\%, 90\%] }}\\
\midrule
Round 1 (90\%)        & \percentage{0.677} & \percentage{0.678} & \percentage{0.790} & \percentage{0.750} & \percentage{0.831} & \percentage{0.759} & \percentage{0.695} & \percentage{0.821} &  \percentage{0.750} \\
Round 2 (90\%)        & \percentage{0.645} & \percentage{0.678} & \percentage{0.810} & \percentage{0.781} & \percentage{0.771} & \percentage{0.775} & \percentage{0.697} & \percentage{0.843} & \percentage{0.750} \\
Round 3 (90\%)        & \percentage{0.710} & \percentage{0.678} & \percentage{0.743} & \percentage{0.760} & \percentage{0.771} & \percentage{0.743} & \percentage{0.677} & \percentage{0.837} & \percentage{0.740} \\
Round 4 (90\%)        & \percentage{0.613} & \percentage{0.670} & \percentage{0.771} & \percentage{0.771} & \percentage{0.754} & \percentage{0.739} & \percentage{0.680} & \percentage{0.821} & \percentage{0.727} \\
Round 0-4             & \percentage{0.677} & \percentage{0.722} & \percentage{0.790} & \percentage{0.760} & \percentage{0.805} & \percentage{0.775} & \percentage{0.720} & \percentage{0.846} & \percentage{0.762} \\
\midrule
\makecell{Lesions mined\\for Round 4}  & \makecell{187\\(3.0\%)} & \makecell{539\\(8.5\%)} & \makecell{682\\(10.8\%)} & \makecell{761\\(12.0\%)} & \makecell{1142\\(18.1\%)} & \makecell{693\\(11.0\%)} & \makecell{768\\(12.2\%)} & \makecell{1545\\(24.5\%)} & - \\
\midrule
\multicolumn{10}{c}{\textbf{Semi-variable Policy [90\%, 90\%, 85\%, 85\%]}}\\
\midrule
Round 1 (90\%)        & \percentage{0.677} & \percentage{0.678} & \percentage{0.790} & \percentage{0.750} & \percentage{0.831} & \percentage{0.759} & \percentage{0.695} & \percentage{0.821} & \percentage{0.750} \\
Round 2 (90\%)        & \percentage{0.645} & \percentage{0.678} & \percentage{0.810} & \percentage{0.781} & \percentage{0.771} & \percentage{0.775} & \percentage{0.697} & \percentage{0.843} & \percentage{0.750} \\
Round 3 (85\%)        & \percentage{0.677} & \percentage{0.721} & \percentage{0.771} & \percentage{0.740} & \percentage{0.767} & \percentage{0.731} & \percentage{0.674} & \percentage{0.796} & \percentage{0.735} \\
Round 4 (85\%)        & \percentage{0.710} & \percentage{0.783} & \percentage{0.829} & \percentage{0.781} & \percentage{0.767} & \percentage{0.735} & \percentage{0.651} & \percentage{0.806} & \percentage{0.758} \\
Round 0-4             & \percentage{0.710} & \percentage{0.748} & \percentage{0.810} & \percentage{0.750} & \percentage{0.810} & \percentage{0.775} & \percentage{0.726} & \percentage{0.823} &  \percentage{0.769} \\
\midrule
\makecell{Lesions mined\\for Round 4}  & \makecell{301\\(2.7\%)} & \makecell{729\\(6.6\%)} & \makecell{1043\\(9.5\%)} & \makecell{1119\\(10.1\%)} & \makecell{1902\\(17.2\%)} & \makecell{1932\\(17.5\%)} & \makecell{1509\\(13.7\%)} & \makecell{2494\\(22.6\%)} & - \\
\midrule
\multicolumn{10}{c}{\textbf{Variable Policy [90\%, 85\%, 80\%, 75\%] }}\\
\midrule
Round 1 (90\%)        & \percentage{0.677} & \percentage{0.678} & \percentage{0.790} & \percentage{0.750} & \percentage{0.831} & \percentage{0.759} & \percentage{0.695} & \percentage{0.821} & \percentage{0.750} \\
Round 2 (85\%)        & \percentage{0.710} & \percentage{0.730} & \percentage{0.829} & \percentage{0.781} & \percentage{0.780} & \percentage{0.771} & \percentage{0.695} & \percentage{0.818} & \percentage{0.764} \\
Round 3 (80\%)        & \percentage{0.710} & \percentage{0.765} & \percentage{0.800} & \percentage{0.729} & \percentage{0.780} & \percentage{0.771} & \percentage{0.646} & \percentage{0.793} & \percentage{0.749} \\
Round 4 (75\%)        & \percentage{0.742} & \percentage{0.730} & \percentage{0.810} & \percentage{0.708} & \percentage{0.809} & \percentage{0.708} & \percentage{0.602} & \percentage{0.808} & \percentage{0.740} \\
Round 0-4             & \percentage{0.774} & \percentage{0.765} & \percentage{0.838} & \percentage{0.760} & \percentage{0.818} & \percentage{0.783} & \percentage{0.720} & \percentage{0.824} & \percentage{0.785} \\
\midrule
\makecell{Lesions mined\\for Round 4}  & \makecell{481\\(2.7\%)} & \makecell{1239\\(6.8\%)} & \makecell{1768\\(9.8\%)} & \makecell{1895\\(10.5\%)} & \makecell{3022\\(16.7\%)} & \makecell{3065\\(16.9\%)} & \makecell{2630\\(14.5\%)} & \makecell{4006\\(22.1\%)} & - \\
\bottomrule
\end{tabular}
\end{adjustbox}
\end{center}
\end{table}

\begin{table}[ht]
\centering\fontsize{10}{12}\selectfont 
\setlength\aboverulesep{0pt}\setlength\belowrulesep{0pt} 
\setlength{\tabcolsep}{7pt} 
\setcellgapes{3pt}\makegapedcells 
\caption{Results of the YOLOX model self-trained with dataset upsampling for different threshold policies. Sensitivities are shown at 4 FP and at 50\% IoU threshold.}
\label{tab:ExpYOLOXUpsampled}
\begin{center}
\begin{adjustbox}{max width=\textwidth, max height=7cm}
\begin{tabular}{@{} c|c|c|c|c|c|c|c|c|c @{}} 
\toprule
Round         & Bone       & Kidney     & Soft Tissue       & Pelvis       & Liver       & Mediastinum       & Abdomen       & Lung      & Mean (95\% CI)\\
\midrule
\multicolumn{10}{c}{\textbf{No Self-Training (Baseline)}}\\
\midrule
Round 0      & \percentage{0.258} & \percentage{0.522} & \percentage{0.476} & \percentage{0.542} & \percentage{0.593} & \percentage{0.506} & \percentage{0.337} & \percentage{0.596} &  \percentage{0.479} \\
\midrule
\makecell{Lesions\\at baseline}  & \makecell{97\\(2.6\%)} & \makecell{195\\(5.2\%)} & \makecell{288\\(7.6\%)} & \makecell{321\\(8.5\%)} & \makecell{426\\(11.3\%)} & \makecell{613\\(16.8\%)} & \makecell{788\\(20.9\%)} & \makecell{1039\\(27.6\%)} & - \\
\midrule
\multicolumn{10}{c}{\textbf{Static Policy [90\%, 90\%, 90\%, 90\%] }}\\
\midrule
Round 1 (90\%)        & \percentage{0.290} & \percentage{0.513} & \percentage{0.543} & \percentage{0.594} & \percentage{0.568} & \percentage{0.530} & \percentage{0.372} & \percentage{0.618} & \percentage{0.504} \\
Round 2 (90\%)        & \percentage{0.258} & \percentage{0.487} & \percentage{0.543} & \percentage{0.542} & \percentage{0.644} & \percentage{0.522} & \percentage{0.409} & \percentage{0.608} & \percentage{0.502} \\
Round 3 (90\%)        & \percentage{0.226} & \percentage{0.452} & \percentage{0.571} & \percentage{0.490} & \percentage{0.581} & \percentage{0.553} & \percentage{0.366} & \percentage{0.611} & \percentage{0.481} \\
Round 4 (90\%)        & \percentage{0.226} & \percentage{0.461} & \percentage{0.467} & \percentage{0.604} & \percentage{0.564} & \percentage{0.502} & \percentage{0.340} & \percentage{0.589} & \percentage{0.469} \\
Round 0-4             & \percentage{0.194} & \percentage{0.470} & \percentage{0.505} & \percentage{0.438} & \percentage{0.530} & \percentage{0.518} & \percentage{0.360} & \percentage{0.586} & \percentage{0.450} \\
\midrule
\makecell{Lesions mined\\for Round 4}  & \makecell{1\\(7.7\%)} & \makecell{0\\(0.0\%)} & \makecell{0\\(0.0\%)} & \makecell{0\\(0.0\%)} & \makecell{1\\(7.7\%)} & \makecell{8\\(61.5\%)} & \makecell{0\\(0.0\%)} & \makecell{3\\(23.1\%)} & - \\
\midrule
\multicolumn{10}{c}{\textbf{Semi-variable Policy [90\%, 90\%, 85\%, 85\%]}}\\
\midrule
Round 1 (90\%)        & \percentage{0.290} & \percentage{0.513} & \percentage{0.543} & \percentage{0.594} & \percentage{0.568} & \percentage{0.530} & \percentage{0.372} & \percentage{0.618} & \percentage{0.504} \\
Round 2 (90\%)        & \percentage{0.258} & \percentage{0.487} & \percentage{0.543} & \percentage{0.542} & \percentage{0.644} & \percentage{0.522} & \percentage{0.409} & \percentage{0.608} & \percentage{0.502} \\
Round 3 (85\%)        & \percentage{0.258} & \percentage{0.522} & \percentage{0.467} & \percentage{0.531} & \percentage{0.530} & \percentage{0.561} & \percentage{0.354} & \percentage{0.602} & \percentage{0.478} \\
Round 4 (85\%)        & \percentage{0.290} & \percentage{0.513} & \percentage{0.457} & \percentage{0.448} & \percentage{0.525} & \percentage{0.569} & \percentage{0.349} & \percentage{0.586} & \percentage{0.467} \\
Round 0-4             & \percentage{0.290} & \percentage{0.478} & \percentage{0.476} & \percentage{0.490} & \percentage{0.551} & \percentage{0.514} & \percentage{0.383} & \percentage{0.580} & \percentage{0.470} \\
\midrule
\makecell{Lesions mined\\for Round 4}  & \makecell{26\\(3.9\%)} & \makecell{26\\(3.9\%)} & \makecell{64\\(9.5\%)} & \makecell{2\\(0.3\%)} & \makecell{201\\(29.9\%)} & \makecell{27\\(4.0\%)} & \makecell{9\\(1.3\%)} & \makecell{318\\(47.3\%)} & - \\
\midrule
\multicolumn{10}{c}{\textbf{Variable Policy [90\%, 85\%, 80\%, 75\%] }}\\
\midrule
Round 1 (90\%)        & \percentage{0.290} & \percentage{0.513} & \percentage{0.543} & \percentage{0.594} & \percentage{0.568} & \percentage{0.530} & \percentage{0.372} & \percentage{0.618} & \percentage{0.504} \\
Round 2 (85\%)        & \percentage{0.258} & \percentage{0.478} & \percentage{0.524} & \percentage{0.552} & \percentage{0.619} & \percentage{0.625} & \percentage{0.427} & \percentage{0.602} & \percentage{0.511} \\
Round 3 (80\%)        & \percentage{0.323} & \percentage{0.478} & \percentage{0.467} & \percentage{0.458} & \percentage{0.513} & \percentage{0.470} & \percentage{0.357} & \percentage{0.530} & \percentage{0.450} \\
Round 4 (75\%)        & \percentage{0.290} & \percentage{0.383} & \percentage{0.448} & \percentage{0.542} & \percentage{0.504} & \percentage{0.494} & \percentage{0.305} & \percentage{0.530} & \percentage{0.437} \\
Round 0-4             & \percentage{0.323} & \percentage{0.557} & \percentage{0.533} & \percentage{0.615} & \percentage{0.585} & \percentage{0.613} & \percentage{0.392} & \percentage{0.573} & \percentage{0.524} \\
\midrule
\makecell{Lesions\\at baseline}  & \makecell{82\\(1.4\%)} & \makecell{339\\(5.8\%)} & \makecell{786\\(13.5\%)} & \makecell{374\\(6.4\%)} & \makecell{940\\(16.1\%)} & \makecell{1041\\(17.9\%)} & \makecell{410\\(7.0\%)} & \makecell{1848\\(31.7\%)} & - \\
\bottomrule
\end{tabular}
\end{adjustbox}
\end{center}
\end{table}

   \begin{figure} [ht]
   \begin{center}
   \begin{tabular}{c} 
   \includegraphics[height=10cm]{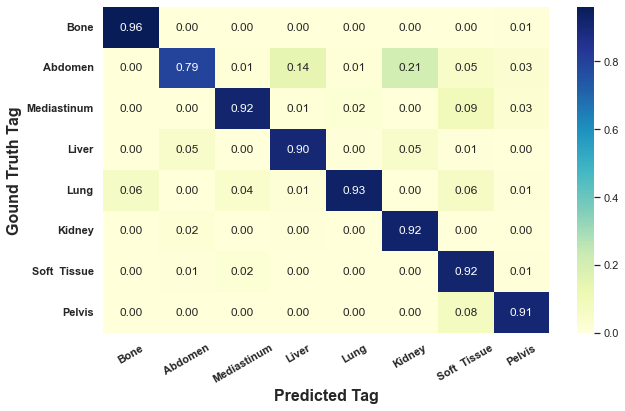}
   \end{tabular}
   \end{center}
   \caption[example] 
   { \label{fig:confusionmat} The confusion matrix shown as heatmap for the VFNet model with upsampled self-training using the variable threshold policy $E_{V}$. The ``abdomen'' lesions are sometimes confused with liver and kidney lesions, possibly due to anatomical proximity.}
   \end{figure} 
\begin{figure} [!h]
\begin{center}
\begin{tabular}{c} 
\includegraphics[width=0.95\columnwidth, height=15cm]{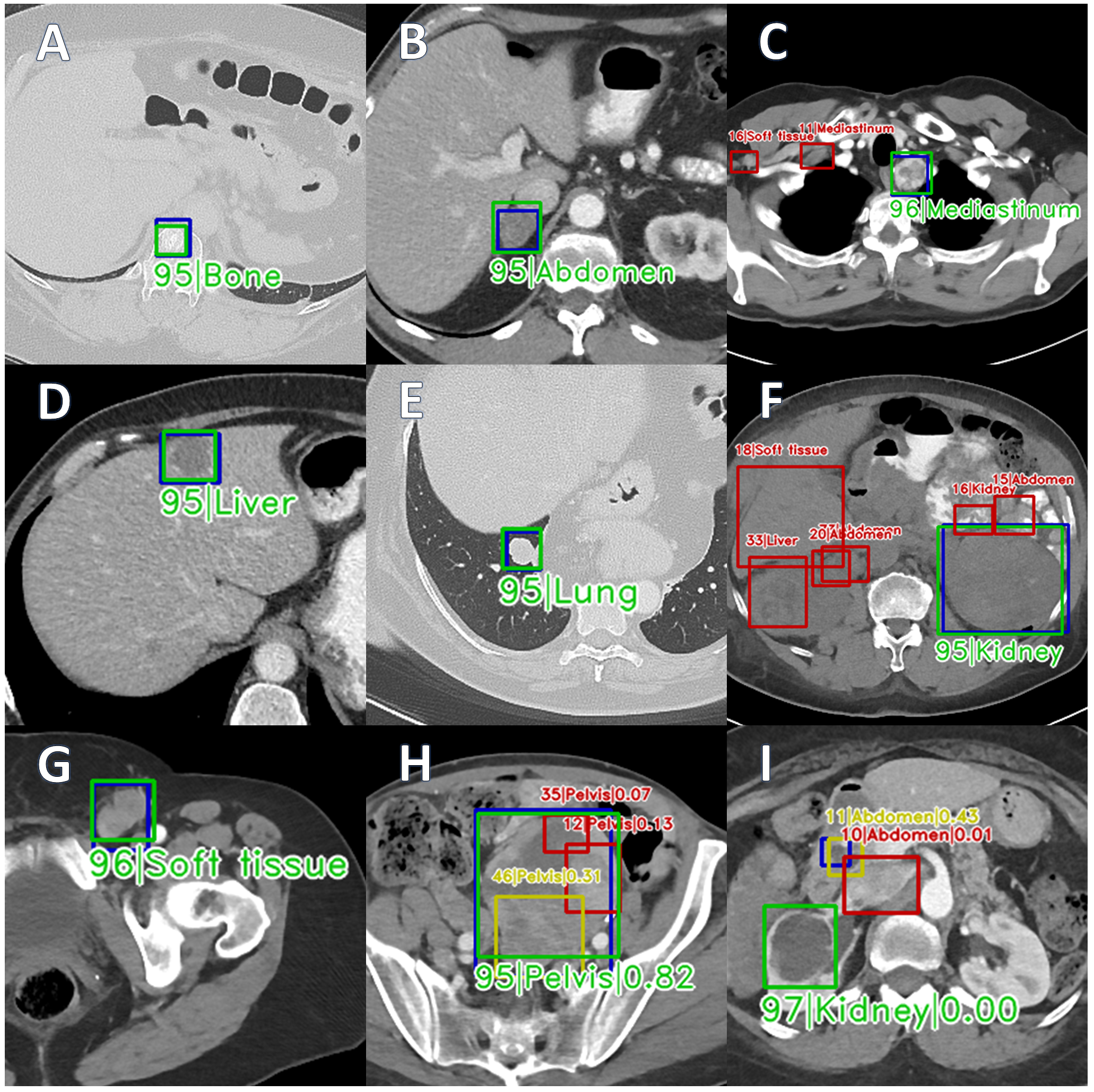}
\end{tabular}
\end{center}
\caption[example] 
{ \label{fig:example} CT slices with predicted lesions to demonstrate the self-training data selection process. The detected lesions are plotted as ``score|lesion type'' for images A-G and ``score|lesion type|IOU with ground truth'' in images H and I. Blue boxes: ground truth. Green boxes: lesions predicted with confidences $\geq$90\%. Yellow boxes: lesions predicted with confidences $<$90\% but with IOU $\geq$30\% with the GT. Red boxes: false positives. Note the GT boxes are from the original DeepLesion training set and were not used in our study. They are plotted only for visualization. \textbf{A-H}: lesions were predicted correctly with the correct lesion type and included in self-training. \textbf{A} is a bone lesion located in the thoracic vertebral body. \textbf{B} is a lesion found at the right adrenal gland adjacent to the liver. \textbf{C} is a left mediastinal lesion, \textbf{D} is a liver mass, and \textbf{E} is a right lung nodule. \textbf{F} shows a large lesion in the left kidney, while there is a possible lesion in the right kidney as well. \textbf{G} is a lesion of the left inguinal lymph node. \textbf{H} show multiple pelvic lesions that were found in the pelvis. \textbf{I} shows an interesting case in which the selected kidney lesion with high confidence is actually abnormal but not labeled in ground truth. On the other hand, the ground truth lesion at the pancreatic head shown in \textbf{I} was detected but not selected (with a confidence score of 11). All images are cropped and zoomed in for better visualization.}
\end{figure} 

\end{document}